**Modeling Chaotic Pedestrian Behavior Using Chaos Indicators and Supervised Learning**

**Md. Muhtashim Shahrier**
Graduate Research Assistant, Department of Civil Engineering
Bangladesh University of Engineering and Technology (BUET), Dhaka, Bangladesh, 1000
Tel: +880-2-55167100, Ext. 7225; Fax: +880-2-58613046; Email: shahriermuhtashim@gmail.com

**Nazmul Haque**
Lecturer, Accident Research Institute (ARI)
Bangladesh University of Engineering and Technology (BUET), Dhaka, Bangladesh, 1000
Tel: +8801673174808, Fax: +880-2-58610081; Email: nhaque@ari.buet.ac.bd

**Md Asif Raihan***
Associate Professor, Accident Research Institute (ARI)
Bangladesh University of Engineering and Technology (BUET), Dhaka, Bangladesh, 1000
Cell: + 8801911142802; Email: raihan@ari.buet.ac.bd

**Md. Hadiuzzaman**
Professor, Department of Civil Engineering
Bangladesh University of Engineering and Technology (BUET), Dhaka, Bangladesh, 1000
Tel: +880-2-55167100, Ext. 7225, Fax: +880-2-58613046; Email: mhadiuzzaman@ce.buet.ac.bd

*Corresponding Author

## ABSTRACT

As cities around the world aim to improve walkability and safety, understanding the irregular and unpredictable nature of pedestrian behavior has become increasingly important. This study introduces a data-driven framework for modeling chaotic pedestrian movement using empirically observed trajectory data and supervised learning. Videos were recorded during both daytime and nighttime conditions to capture pedestrian dynamics under varying ambient and traffic contexts. Pedestrian trajectories were extracted through computer vision techniques, and behavioral chaos was quantified using four chaos metrics: Approximate Entropy and Lyapunov Exponent, each computed for both velocity and direction change. A Principal Component Analysis (PCA) was then applied to consolidate these indicators into a unified chaos score. A comprehensive set of individual, group-level, and contextual traffic features was engineered and used to train Random Forest and CatBoost regression models. CatBoost models consistently achieved superior performance. The best daytime PCA-based CatBoost model reached an $R^2$ of 0.8319, while the nighttime PCA-based CatBoost model attained an $R^2$ of 0.8574. SHAP analysis highlighted that features such as distance travel, movement duration, and speed variability were robust contributors to chaotic behavior. The proposed framework enables practitioners to quantify and anticipate behavioral instability in real-world settings. Planners and engineers can use chaos scores to identify high-risk pedestrian zones, apprise infrastructure improvements, and calibrate realistic microsimulation models. The approach also supports adaptive risk assessment in automated vehicle systems by capturing short-term motion unpredictability grounded in observable, interpretable features.

# INTRODUCTION

Understanding pedestrian behavior remains a fundamental concern in transportation research, particularly as cities worldwide prioritize walkability, safety, and multimodal mobility. Pedestrians being the most vulnerable road users, often move in areas with little structure, loose rule enforcement, and constantly changing interactions with vehicles and other people. Such conditions are common not only in congested urban centers but also in informal crossings, shared spaces, transit terminals, and festival zones. In these situations, how pedestrians move can change a lot depending on nearby people, space limits, and the surroundings. Pedestrian walking behavior is predominantly chaotic in nature (*1*). Standard modeling frameworks that treat pedestrian flow as a deterministic or uniformly stochastic process often fail to capture the full range of real-world behavior.

Modeling pedestrian behavior has traditionally relied on flow-level metrics such as average speed, density, and delay, or on agent-based simulations grounded in social force or cellular automata models. Early foundational work by Helbing and Molnár formalized the use of social force models to simulate pedestrian interactions, where individual behavior is shaped by proximity to other agents and environmental boundaries (*2*). Traditional pedestrian modeling has often utilized cellular automata (CA) approaches to simulate crowd dynamics and evacuation scenarios (*3*, *4*). Recent advancements have expanded these models to account for pedestrian-vehicle interactions, highlighting the complexities of mixed traffic flow (*5*). Additionally, Markovian models have been proposed to simulate pedestrian behavior and predict car-to-pedestrian accidents, using discrete states like standing, walking, jogging, and running to model pedestrian movement (*6*). Machine learning techniques, particularly deep learning, have also been employed to predict pedestrian behavior in urban scenarios, offering superior performance over traditional models in complex environments (*7*).

Chaos theory has gained attention in transportation engineering for its ability to capture complex, nonlinear system behavior. Narh et al. emphasized its suitability for urban transportation control, arguing that the increasing availability of high-resolution temporal and spatial data makes it a promising tool for detecting chaotic patterns in transportation systems (*8*). Frazier and Kockelman explored the application of chaos theory to short-term forecasting, developing a model for traffic prediction in California based on chaotic dynamics (*9*). Adewumi et al. proposed a method for urban traffic flow prediction using nonlinear time series modeling grounded in chaos theory (*10*). Xu and Gao developed a two-stage flow prediction framework combining the gravity model with user equilibrium principles (*11*).

While chaos theory has been explored in traffic flow and demand modeling, its application to pedestrian dynamics remains comparatively limited, though emerging research suggests its significant potential. Several studies have demonstrated that pedestrian motion, particularly in dense or unregulated environments, exhibits chaotic properties rather than mere randomness. Liu and Li introduced the concept of a pedestrian flow Reynolds number and employed nonlinear time-series analysis, including Lyapunov exponents, to understand the pedestrian movement under crowd pressure and directional fluctuation (*12*). Qi et al. applied techniques such as the Largest Lyapunov Exponent and correlation dimension to pedestrian flow data to validate the presence of chaos, with the aim of informing future interactions with self-driving vehicles (*13*). Saeedi and Rassafi modeled pedestrian crossing behavior using the Helbing social force framework with synthetic data and performed sensitivity analysis to detect the onset of chaos under varying crowd conditions (*14*). While these studies confirm the existence of chaotic behavior in pedestrian, they primarily focus on detecting or describing the phenomenon. The next step lies in building predictive models that link observable features to quantified measures of chaos, allowing practitioners to anticipate and manage behavioral unpredictability in real time.

Despite growing evidence that pedestrian movement can exhibit chaotic dynamics, the existing body of research has largely remained descriptive or simulation-based. Most studies stop at verifying the presence of chaos using time-series analysis or theoretical modeling, without progressing toward data-driven prediction or real-world application. There is currently a lack of studies that quantify chaos at the individual pedestrian level using empirically observed trajectories and then use those quantified values as targets in supervised learning models. No established framework exists that translates this behavioral instability into a measurable score derived from interpretable features such as walking speed, stop-go behavior, group dynamics, or surrounding traffic context. While machine learning has been applied in pedestrian trajectory prediction, it has not been leveraged to predict behavioral unpredictability or chaotic tendencies. This gap not only limits the practical use of chaos theory in transportation engineering but also leaves unaddressed the potential for real-time risk assessment, simulation realism, or adaptive infrastructure planning based on behavioral instability.

This study proposes a data-driven framework for modeling chaotic pedestrian behavior using real-world trajectory data captured during both day and night. After filtering out short or noisy trajectories, a set of behavioral and contextual features was computed for each pedestrian, including walking patterns, path efficiency, group density, and surrounding vehicle dynamics. The analysis was spatially separated by roadway side to isolate local interactions between pedestrians and traffic. Vehicle-related features were aggregated in temporal bins and assigned to pedestrians based on their presence window. Chaos was quantified using four metrics: Lyapunov Exponent and Approximate Entropy for both directional change and change in velocity. These formed the target variables for supervised learning. Predictive models were trained individually for each chaos metric using Random Forest and CatBoost regressors. Additionally, Principal Component Analysis was applied to the chaos scores to derive a composite measure, which was also modeled to explore its relation with observable pedestrian and environmental features.

The ability to quantify and predict chaotic pedestrian behavior has direct implications for a range of transportation applications. Urban planners and traffic engineers can use such models to identify locations or times where pedestrian behavior becomes highly unstable, informing the design of crossings, barriers, or signalization strategies. Simulation developers can incorporate chaos-based variability to enhance the realism of pedestrian microsimulation, especially in shared spaces or low-control environments. For safety analysts, chaos scores can serve as proactive risk indicators, identifying areas where erratic movement increases collision likelihood. In the context of connected and automated vehicles, understanding pedestrian behavioral instability can support the development of more adaptive, context-aware control systems. Additionally, this approach offers a pathway for integrating behavioral complexity into pedestrian impact assessments, infrastructure audits, and real-time monitoring tools, making it valuable for both long-term planning and operational decision-making.

The remainder of this paper is organized as follows. The next section outlines the overall methodology of the study, including the theoretical basis for the chaos metrics and the modeling framework. This is followed by a detailed description of the data collection process, trajectory extraction, and feature engineering procedures. The model development section presents the machine learning pipeline. Finally, the paper concludes with a discussion of results.

## METHODOLOGY

This study models chaotic pedestrian behavior by analyzing individual-level trajectory patterns and predicting the resulting degree of motion irregularity using a combination of behavioral, contextual, and group-based features.

Trajectory data were collected from videos recorded on a national highway under both daytime and nighttime conditions. These recordings were captured in free-flowing traffic segments away from intersections. From the video data, pedestrian trajectories were extracted using computer vision algorithms, generating frame-wise coordinates of each individual's movement over time. Only those trajectories where a pedestrian remained in the frame for longer than four seconds were retained for analysis, in order to ensure sufficient temporal resolution for dynamic modeling. Camera calibration was conducted to convert pixel coordinates to real-world distances, using a quadratic mapping derived from scene geometry. The image was conceptually divided into two lateral regions corresponding to each side of the roadway, and pedestrians were assigned to a side based on their entry location. This partitioning ensured that each pedestrian was influenced only by the traffic dynamics occurring in their immediate context.

### Feature Engineering

Once the raw trajectories were filtered and calibrated, a series of explanatory variables were derived for each pedestrian. These included movement-based features, group-level attributes, and environmental indicators related to surrounding vehicle behavior. Movement duration and distance traveled were directly computed from the start and end frames and the spatial displacement of each pedestrian. Mean speed was defined as the average Euclidean distance covered per frame, while speed variation captured the standard deviation across the pedestrian's speed profile. A stop-and-go duration variable was created where the pedestrian's speed dropped below 0.2 meters per second, indicating momentary pauses or hesitation during movement Entry angle was defined as the orientation of the trajectory vector during the initial segment of movement, measured in radians. It captured the direction from which the pedestrian entered the scene. A path efficiency score was computed to quantify trajectory directness. This metric reflects the proportion of the actual path that deviated from a hypothetical straight-line trajectory between the pedestrian's start and end locations.

Group-level features included the average pedestrian density within a defined spatial radius while the focal pedestrian was present in the scene. Density was calculated as the average number of other pedestrians present per square meter within a defined spatial radius across the full trajectory duration. Environmental vehicle-related variables were aggregated from macroscopic traffic metrics using one-minute bins. Within each bin, vehicle flow was computed as vehicles per hour per lane, and mean speed and speed variation were derived from the detected vehicle population. Vehicle composition was defined as the proportion of light, medium, and heavy vehicles present, determined by class labels generated during object detection. For each pedestrian, the values from the bin containing their highest residence time were assigned as contextual indicators. This ensured alignment between pedestrian behavior and concurrent traffic conditions.

**Chaos Metric Computation**

The chaos metrics used as target variables in this study quantify the complexity and stability of pedestrian trajectories. These metrics include Lyapunov Exponents, which measure sensitivity to initial conditions and reflect dynamic instability, and Approximate Entropy, which captures the irregularity and unpredictability of time series. Both sets of indicators were computed separately for velocity and directional behavior, resulting in four chaos variables per pedestrian.

This study selected Lyapunov Exponent and Approximate Entropy as the primary indicators to comprehensively quantify chaos in pedestrian movement. These two measures capture complementary aspects of complex system behavior. The Lyapunov Exponent, grounded in dynamical systems theory, reflects sensitivity to initial conditions and quantifies how rapidly nearby trajectories diverge, making it suitable for assessing long-term instability. In contrast, Approximate Entropy evaluates the short-term irregularity of time series data, measuring the predictability of temporal patterns within a finite observation window. By combining these two indicators, the study aimed to characterize both the structural instability and statistical unpredictability inherent in naturalistic pedestrian trajectories. This dual-metric approach also enhances robustness, allowing the modeling framework to capture diverse manifestations of chaotic behavior in real-world settings.

To further capture the multi-dimensional nature of pedestrian irregularity, each chaos metric was computed separately for two aspects of movement: velocity and direction change. Velocity-based metrics reflect temporal instability, identifying patterns such as inconsistent walking speed, abrupt stops, or acceleration spikes. Directional metrics, on the other hand, quantify spatial irregularity such as swerving, zigzag motion, or hesitancy in path selection. These two behavioral dimensions often manifest independently in real-world scenarios, particularly in unstructured pedestrian environments. By analyzing both, the study ensures a more holistic representation of behavioral chaos that accounts for instability in both how fast and in what manner a pedestrian move.

The Lyapunov exponent is a fundamental concept in dynamical systems and chaos theory, used to measure the sensitivity of a system to initial conditions. It provides a quantitative assessment of how small perturbations in the system's state evolve over time, offering insight into stability, predictability, and chaotic behavior (*15*). The Lyapunov exponent determines whether small differences in an initial state amplify or diminish over time. It is defined as **Equation 1:**

$$\lambda = \lim_{t\to\infty} \frac{1}{t} \ln \frac{||\delta X(t)||}{||\delta X(0)||} \quad (1)$$

δX(0) represents an initial small perturbation in the system, δX(t) is the perturbation after time t, and λ (the Lyapunov exponent) quantifies the rate at which two nearby trajectories diverge over time. A Lyapunov Exponent greater than zero indicates chaotic behavior, while values close to zero suggest borderline dynamics, and negative values reflect stable, converging motion. This interpretation allows the metric to distinguish between unstable and orderly pedestrian trajectories. In this study, the Lyapunov Exponent was computed separately for the pedestrian's direction and velocity time series using the Rosenstein method (*16*). For each pedestrian, the trajectory was reconstructed in phase space using a time-delay embedding with fixed dimension and lag. The average divergence between neighboring trajectories was tracked over time, and the slope of the resulting divergence curve was used to estimate the Lyapunov Exponent. This process was implemented using custom MATLAB code.

Approximate Entropy (ApEn) is a statistical measure that evaluates the regularity and complexity of time series data. It quantifies the likelihood that similar patterns of observations remain similar in subsequent time steps (*17*). A lower ApEn value indicates more structured, predictable behavior, whereas higher values signify increased irregularity and randomness. Mathematically, Approximate Entropy is defined as **Equation 2:**

$$ApEn(m, r, N) = \Phi^m(r) - \Phi^{m+1}(r) \quad (2)$$

Where m is the embedding dimension (number of previous points considered), r is the tolerance (threshold for similarity), N is the total number of data points, and $\Phi^m(r)$ is a measure of the probability that two sequences of length m remain similar when extended to m+1 where $\Phi^m(r)$ is defined as **Equation 3:**

$$\Phi^m(r) = \frac{1}{N-m+1} \sum_{i=1}^{N-m+1} \ln C_i^m(r) \quad (3)$$

Where $C_i^m(r)$ is the relative frequency of template vectors being similar (within tolerance r) to vector *X(i)* and *X(i)* represents a sequence of m consecutive data points from the original time series. Approximate Entropy typically ranges between 0 and 2 for normalized time series, where lower values correspond to highly regular, predictable behavior, and higher values reflect greater randomness and complexity in the trajectory. In this study, Approximate Entropy was calculated for two derived time series: pedestrian velocity and direction change. Velocity was defined as the Euclidean distance between consecutive positions divided by frame time, and direction change was computed from the angular change in movement vectors across frames. Both series were processed using custom Python functions, with embedding dimension m set to 2 and tolerance r set to 20% of the standard deviation of the input series.

**Composite Chaos Score via Principal Component Analysis**

Once all four chaos indicators were computed, they were assembled into the dataset alongside the explanatory features. As a final preprocessing step, Principal Component Analysis (PCA) was applied to the chaos indicators to extract a single composite variable that captured overall motion irregularity. PCA is a dimensionality reduction technique that transforms correlated variables into orthogonal components, ranked by the variance they explain (*18*). Before applying PCA, the chaos metrics were standardized to have zero mean and unit variance. A one-component PCA model was then trained using the PCA module in Python's scikit-learn library (*19*).

To interpret the composite chaos score, the loadings of each metric on the first principal component were examined. Approximate Entropy of Velocity and Direction Change showed the highest contributions, with weights of 0.6958 and 0.6995 respectively, indicating that irregular fluctuations in speed and directional changes were the dominant sources of behavioral instability in the dataset. Lyapunov-based metrics had comparatively lower influence, with Lyapunov Velocity change contributing 0.1199 and Lyapunov Direction Change showing a slight negative weight (–0.1099), suggesting limited impact on the overall motion irregularity captured by the PCA. The first principal component explained 40.63% of the total variance in the standardized chaos metrics, consolidating multiple dimensions of unpredictability into a single interpretable score.

PCA was applied for two reasons. First, while the four chaos metrics capture distinct aspects of behavioral instability, they are inherently correlated due to their shared dependence on the underlying pedestrian trajectory. Modeling them independently could lead to redundant or noisy representations. By using PCA, these interdependencies were consolidated into a single composite variable that retained the dominant variance structure of the original metrics. Second, a unified chaos score enables clearer interpretation and more stable model training, especially when evaluating feature importance. This approach not only reduces dimensionality but also improves the robustness of downstream prediction and interpretability, allowing the model to focus on generalized patterns of chaos rather than metric-specific noise.

**Supervised Learning Models**

Random Forest regression was used to model the relationship between explanatory features and each chaos indicator. Random Forest (RF) is an ensemble-based regression method that constructs a collection of decision trees to reduce both bias and variance in prediction (20). The algorithm operates by building multiple tree-based models on different bootstrapped subsets of the training data and then averaging their outputs to generate a final prediction. Each individual tree partitions the input space using recursive binary splits on the explanatory variables, producing a piecewise approximation of the target function. By aggregating the results across multiple trees, the ensemble is able to capture non-linear relationships while mitigating the risk of overfitting.

In this study, the Random Forest algorithm was used to model the relationship between pedestrian-level features and each chaos metric individually. The model was implemented in Python with 100 estimators. The input dataset included behavioral, group-level, and contextual variables, while the pedestrian ID and all chaos metrics were excluded from the feature matrix during training. For each modeling run, the target variable was one of the four chaos indicators: Approximate Entropy of velocity, Approximate Entropy of direction change, Lyapunov Exponent of velocity, or Lyapunov Exponent of position. In addition to these, the PCA-derived composite score capturing overall motion irregularity was also modeled as a separate target. The dataset was partitioned into training and testing subsets using an 80–20 split, and a fixed random seed was applied to ensure reproducibility of results.

The RF algorithm trained each decision tree on a random sample of the training data drawn with replacement; a process known as bootstrapping. At each node of a tree, a random subset of features was selected, and the best variable among them was chosen to perform the split. This strategy of random feature selection prevented dominance by strong predictors and encouraged model diversity across trees. The out-of-bag (OOB) data, comprising the unused samples in each bootstrap draw, were reserved internally by the algorithm to support unbiased estimation and reduce generalization error.

CatBoost is a gradient boosting algorithm that builds an ensemble of decision trees in a sequential manner, where each tree attempts to correct the errors made by its predecessors (21). Unlike traditional boosting methods, CatBoost introduces advanced techniques to handle categorical variables natively, reduce prediction shift, and improve generalization through efficient gradient-based optimization. The algorithm constructs trees using symmetric tree structures and applies ordered boosting to prevent overfitting, particularly in cases with small to medium-sized datasets.

In this study, CatBoost regression was employed to model the relationship between observable pedestrian and contextual features and each of the four chaos indicators and the PCA derived score. The model was implemented using Python. The dataset was split into training and testing subsets using an 80–20 ratio, with a fixed random seed to ensure consistency. The CatBoost model was trained using 1,000 boosting iterations with a learning rate of 0.05 and a maximum tree depth of 6. The model was configured to display verbose output every 100 iterations to monitor convergence behavior. By leveraging gradient-based updates and its inbuilt handling of numerical feature distributions, CatBoost was able to capture both linear and complex non-linear relationships between the input features and the chaos scores.

All the models were evaluated based on $R^2$ values and Root Mean Square Error (RMSE) values.

# DATA COLLECTION AND DATASET PREPARATION

This study utilized naturalistic traffic footage recorded on a segment of the National Highway located on the outskirts of Dhaka, Bangladesh. This site was selected due to its exposure to both high pedestrian activity and traffic flow. To capture behavioral variations under different ambient conditions, separate recordings were conducted during both daytime and nighttime periods from the same camera location and angle.

For object detection, the approach varied by lighting condition. During the daytime, pedestrians and vehicles were detected using the DEEGITS model—a deep learning framework developed for accurate detection in complex mixed-traffic scenarios. The model is optimized for the Bangladeshi traffic context and has been validated on local datasets (*22*). For nighttime videos, a separate object detection model, developed for pedestrian and vehicle detection under low-light conditions, was used to address challenges such as poor illumination, headlight glare, and reduced contrast (*23*). Both models produced frame-level bounding boxes and class labels.

Once detection was completed, DeepSORT, a real-time multiple objects tracking algorithm (*24*), was applied to generate continuous trajectory data for each pedestrian across consecutive frames. The tracker assigned unique IDs to each pedestrian and produced frame-wise spatial coordinates, enabling analysis of movement patterns over time. A unified trajectory file was then compiled containing pedestrian ID, class label, frame number, and (x, y) coordinates in image space.

A filtering step was applied to ensure that the extracted trajectories were of sufficient length for meaningful analysis Only pedestrian trajectories lasting longer than four seconds were retained. This threshold was chosen to allow for reliable estimation of dynamic features such as velocity, path tortuosity, and behavioral entropy.

To enable the calculation of physical quantities, pixel-level coordinates were transformed into real-world measurements using a camera calibration process. A quadratic polynomial mapping was established between pixel displacement and physical distance, based on a set of ground reference measurements captured at the scene. This calibration allowed for accurate computation of variables that depend on metric-scale data.

For each retained pedestrian, a comprehensive set of explanatory variables was calculated. In parallel, four chaos-related metrics were computed: Approximate Entropy of velocity and direction change, and Lyapunov Exponents based on positional and velocity trajectories. In addition to the individual metrics, a composite chaos score was derived using Principal Component Analysis (PCA) applied to the standardized chaos indicators.

A cleaning step was performed to eliminate physically implausible entries caused by detection errors or tracking noise. For each variable, acceptable ranges were established based on domain knowledge and exploratory data analysis. Pedestrian instances exhibiting values outside these thresholds for any of the computed variables were excluded from the final dataset. This conservative filtering step was necessary to reduce noise and improve model reliability.

# MODEL DEVELOPMENT

To predict the degree of chaos in pedestrian behavior, multiple supervised regression models were trained using the full set of explanatory variables. These included individual-level motion features, group-based attributes, and contextual traffic indicators. The target variables were the four chaos metrics - Approximate Entropy (velocity and direction change) and Lyapunov Exponents (velocity and direction change), as well as a composite PCA-based score.

All models shared a common pipeline: the dataset was split into training and test subsets using an 80–20 ratio with a fixed random seed for reproducibility. Each model was trained separately for each chaos indicator. Evaluation was based on the $R^2$ score, and Root Mean Squared Error (RMSE) calculated on the test set.

Random Forest (RF) was selected for its ability to handle nonlinear interactions and resist overfitting. The model ensemble consisted of 100 decision trees trained on bootstrapped samples of the training data. Feature selection was randomized at each node to promote diversity across trees.

CatBoost, a gradient boosting algorithm, was employed as a complementary approach. Known for its high performance on tabular data, CatBoost used symmetric tree structures and ordered boosting to reduce overfitting. Models were trained with 1,000 boosting rounds, a learning rate of 0.05, and maximum depth of 6.

To interpret model predictions, SHAP (SHapley Additive exPlanations) values were computed. SHAP analysis identified the most influential features driving each chaos prediction, revealing patterns in how behavioral or contextual factors contributed to motion irregularity.

## RESULTS AND DISCUSSION

This section presents the predictive performance of machine learning models developed to estimate chaotic pedestrian behavior, as captured by both individual chaos indicators and a composite Principal Component Analysis (PCA)-based target. Two different time periods, daytime and nighttime were analyzed independently using Random Forest and CatBoost regressors. Performance was assessed using the coefficient of determination ($R^2$) and the root mean squared error (RMSE). These metrics were selected for their complementary interpretability, where $R^2$ quantifies the proportion of variance explained by the model and RMSE captures the average prediction error in the same units as the target variable.

The coefficient of determination, $R^2$, is defined as **Equation 3:**

$$R^2 = 1 - \frac{\sum_{i=1}^{n}(y_i - \hat{y}_i)^2}{\sum_{i=1}^{n}(y_i - \bar{y})^2} \quad (3)$$

where $y_i$ denotes the actual target value, $\hat{y}_i$ represents the predicted value from the model, $\bar{y}$ is the mean of all actual target values, and n is the number of observations. A higher $R^2$ indicates that a larger proportion of the variability in the target is captured by the model. $R^2$ values can range from negative values (indicating extremely poor performance) up to 1.0 (indicating perfect prediction).

The root mean squared error (RMSE) provides an absolute measure of model error and is expressed as **Equation 4:**

$$RMSE = \sqrt{\frac{\sum_{i=1}^{n}(y_i - \hat{y}_i)^2}{n}} \quad (4)$$

This metric measures the average magnitude of the prediction errors, and lower values suggest that the model is generating predictions that are closer to the actual observations. Unlike $R^2$, which is scale-free, RMSE is measured in the same units as the target variable, making it particularly useful for quantifying the magnitude of predictive deviations.

All model configurations were tested across all chaos indicators and regression algorithms. Only the top-performing models are presented in **Table 1** and **Table 2**. Initial screening revealed that models trained on Lyapunov-based targets consistently performed poorly, with $R^2$ values falling below 0.20 regardless of algorithm choice. These configurations were therefore excluded from detailed evaluation due to their limited explanatory ability. Among the remaining combinations, the majority of models predicting Approximate Entropy metrics produced moderate performance, typically with $R^2$ values ranging from 0.70 to 0.80. From this subset, the three most accurate models for each time period, based on $R^2$ and RMSE, were selected as key performing models. This filtering strategy ensured that the final interpretability analysis was grounded in models with both statistical reliability and practical relevance.

For daytime (**Table 1**), the CatBoost model predicting Approximate Entropy of Direction Change achieved the highest $R^2$ value of 0.8319 and the lowest RMSE of 0.1102. However, a separate CatBoost model trained on the PCA-compressed chaos score achieved an $R^2$ of 0.8131, which, while slightly lower in predictive strength, offers a more comprehensive view of motion irregularity by capturing shared variance across multiple chaos indicators. Therefore, the PCA-based CatBoost model was selected as the representative model for interpretability analysis in the daytime setting. The Random Forest model, while still effective, showed marginally lower $R^2$ values, potentially due to its higher model variance and limited capacity to capture subtle feature interactions in high-dimensional spaces. In contrast, CatBoost's gradient boosting framework allowed it to reduce both bias and variance through sequential tree refinement, making it better suited for modeling the complex, interdependent relationships in pedestrian behavior data.

**TABLE 1 Key Model Performance for Daytime**

| Target variable | Model type | $R^2$ | RMSE |
|---|---|---|---|
| Approximate Entropy Direction Change | Random Forest | 0.8203 | 0.1139 |
| Approximate Entropy Direction Change | CatBoost Regressor | 0.8319 | 0.1102 |
| PCA | CatBoost Regressor | 0.8131 | 0.5709 |

During nighttime (Table 2), all models exhibited improved predictive power. The CatBoost model predicting Approximate Entropy of Direction Change achieved the highest $R^2$ of 0.8726 and the lowest RMSE of 0.1199. The PCA-based CatBoost model was again selected as the representative model due to its capacity to encapsulate multi-dimensional chaos indicators in a single interpretable output. This model achieved an $R^2$ of 0.8574 and an RMSE of 0.4803, reflecting strong predictive performance. As observed earlier, the Random Forest models performed slightly worse, likely due to their limitations in modeling complex feature interactions and handling multicollinearity in high-dimensional input spaces.

**TABLE 2 Key Model Performance for Nighttime**

| **Target variable** | **Model type** | **$R^2$** | **RMSE** |
|---|---|---|---|
| Approximate Entropy Direction Change | Random Forest | 0.8585 | 0.1264 |
| Approximate Entropy Direction Change | CatBoost Regressor | 0.8726 | 0.1199 |
| PCA | CatBoost Regressor | 0.8574 | 0.4803 |

Although the individual Approximate Entropy models yielded slightly higher $R^2$ values, the PCA-based models were selected for interpretation because they consolidate multiple chaos indicators into a single continuous target. This consolidation reduces noise from indicator-specific variance and facilitates a clearer and more stable interpretability analysis. The PCA target used in the study incorporated four metrics, Approximate Entropy of velocity and direction change, along with Lyapunov Exponents derived from both positional and velocity data and was constructed after standardizing all metrics and retaining the first principal component.

An inspection of the PCA component loadings revealed that Approximate Entropy of velocity and direction change were the most influential contributors to the composite chaos score, with both exhibiting weights close to 0.70. In contrast, the Lyapunov-based metrics showed minimal impact, with one contributing slightly negatively. These results indicate that the principal source of behavioral instability in the dataset was short-term irregularity in pedestrian motion, rather than long-range divergence or sensitivity to initial conditions. The first component accounted for 40.63% of the total variance, which, while moderate, effectively captured the dominant structure across the four chaos indicators. This reinforces the decision to use the PCA-derived score as a unified, interpretable measure of pedestrian behavioral chaos in naturalistic settings.

To further illustrate how the composite chaos score captures real behavioral differences, two example pedestrians from the dataset were examined based on their PCA values. These instances represent the extremes of the chaos spectrum and provide interpretability at the individual level.

A trajectory corresponding to a high composite chaos score (PCA = 2.28) was characterized by highly irregular motion. The pedestrian exhibited Approximate Entropy values of 1.63 in velocity and 0.85 in direction change—both near the upper bound of the dataset. These values reflect frequent fluctuations in speed and rapid directional shifts within a short time window. Interestingly, both Lyapunov Exponents were zero, suggesting that the chaotic behavior stemmed primarily from short-term unpredictability rather than dynamic instability. This example reinforces the role of Approximate Entropy as the dominant contributor to pedestrian chaos in naturalistic settings.

Conversely, a sample with a low chaos score (PCA = –2.04) showed highly regular behavior. The pedestrian had very low Approximate Entropy values in both velocity (0.28) and direction change (0.16), indicating consistent speed and a straight trajectory. Lyapunov Exponents were also minimal, with small positive values that suggest no meaningful divergence from initial conditions. This trajectory likely represents a stable, uninterrupted crossing event, validating the model's ability to identify low-chaos behavior patterns when movement is orderly and predictable.

The relationship between the predicted and observed values for the selected PCA-based CatBoost models is shown in Figures 6 and 7. Both models demonstrate strong alignment between predicted and actual chaos scores, with a tight distribution around the 45-degree reference line. The daytime model (**Figure 1a**) shows a more compact spread, while the nighttime model (**Figure 1b**) exhibits slightly greater dispersion but also covers a wider score range, reflecting greater behavioral variability under reduced lighting conditions.

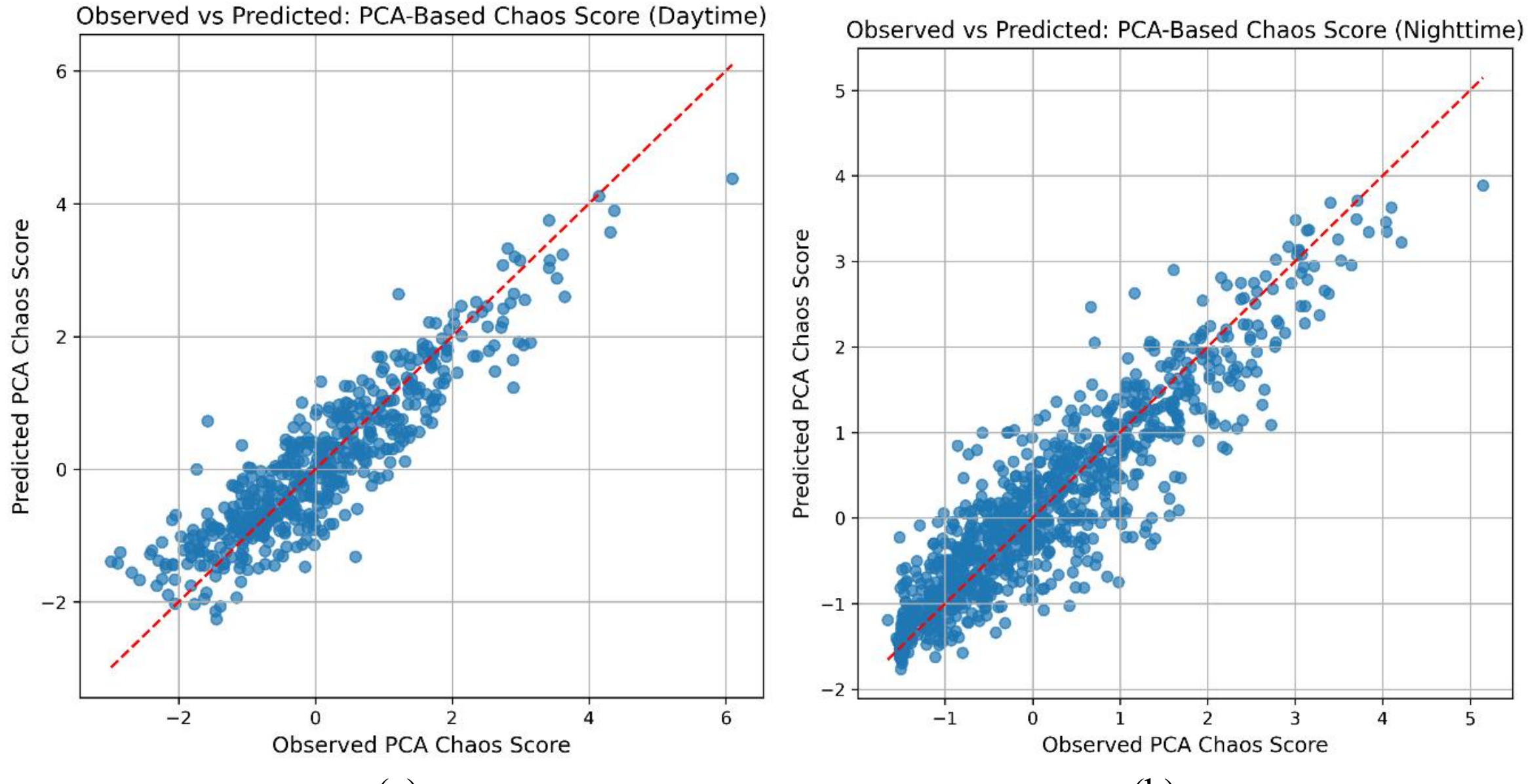

(a) (b)

**Figure 1: Observed vs. predicted PCA-based chaos score for CatBoost model (a) Daytime (b) Nighttime. Dashed red line represents the 1:1 reference line**

Feature contributions to the PCA-based CatBoost models were further examined using SHAP (SHapley Additive exPlanations) values. These values provide a unified framework for quantifying how much each feature contributed to the prediction for a given observation. The beeswarm plots generated from SHAP analysis for both daytime and nighttime illustrated the direction and magnitude of impact for each variable.

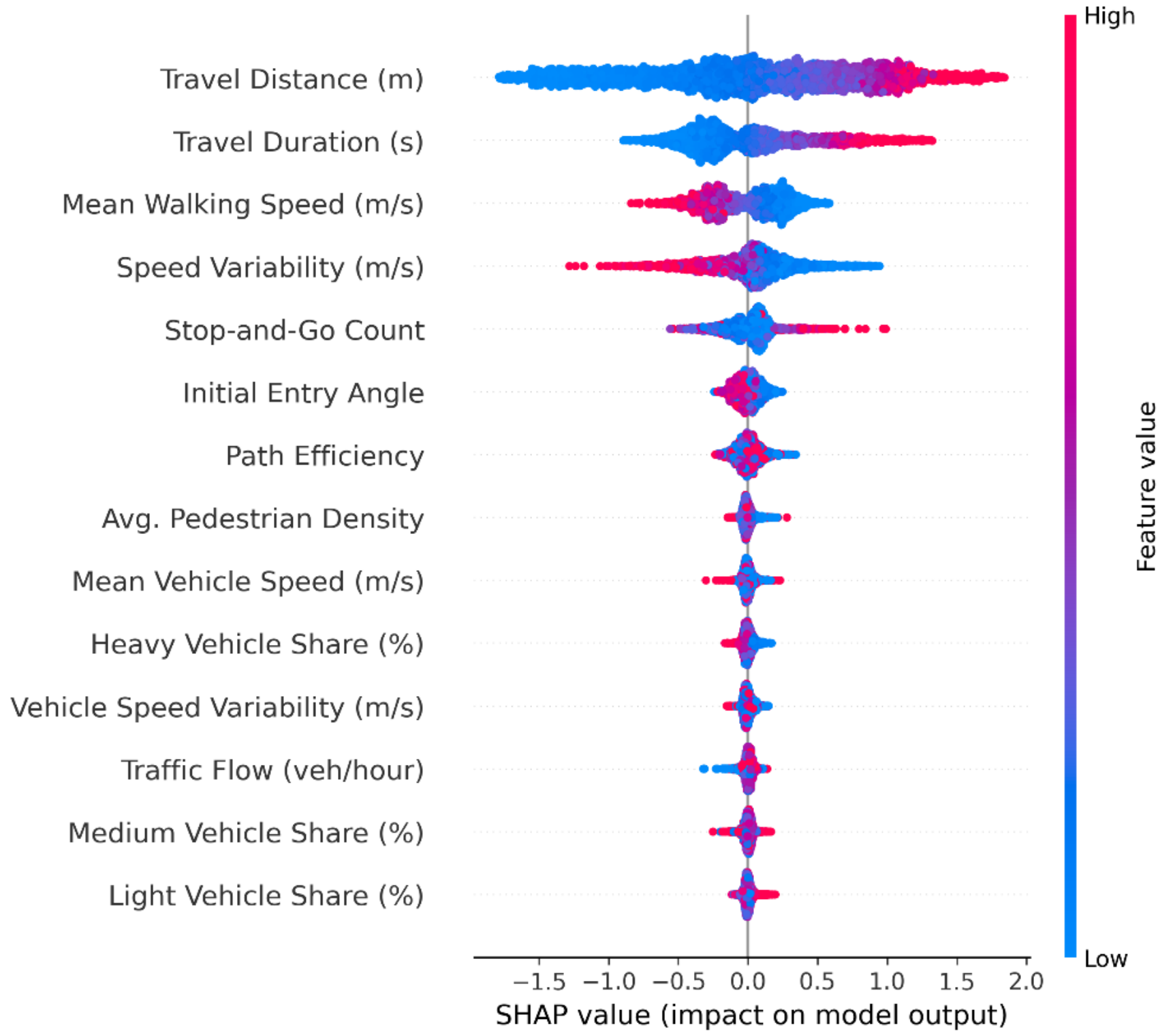

**Figure 2 SHAP Beeswarm Plot for Daytime PCA-based CatBoost Model. Color Represents Feature Value from Low (blue) to High (red)**

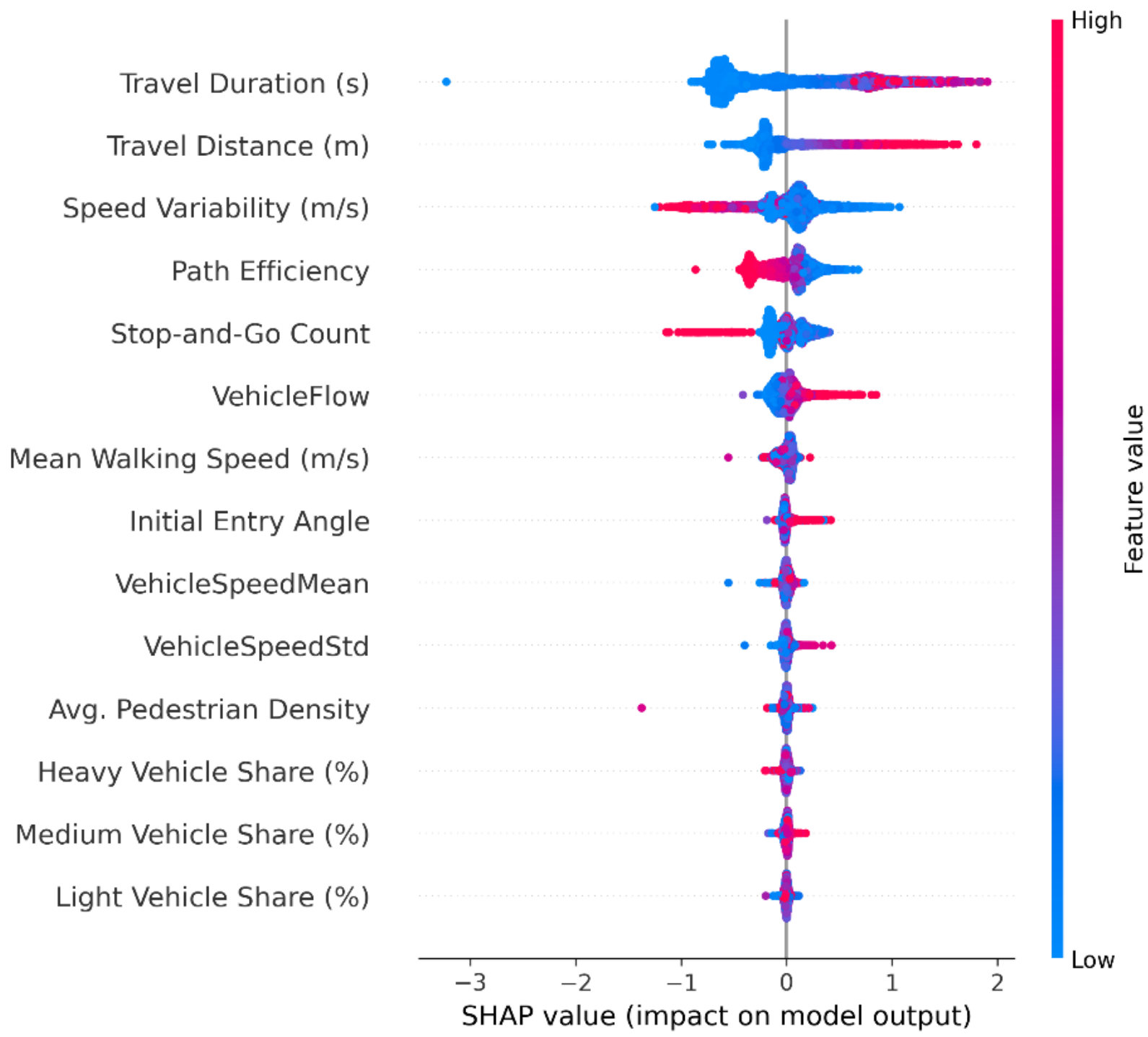

**Figure 3 SHAP Beeswarm Plot for Nighttime PCA-based CatBoost Model. Color Represents Feature Value from Low (blue) to High (red)**

In the daytime model (**Figure 2**), the most influential feature was Travel Distance, followed by Travel Duration. Pedestrians covering greater distances or remaining in the scene for longer durations were more likely to be assigned higher chaos scores. These patterns likely reflect indirect or circuitous trajectories, as well as time-consuming interactions with traffic or other pedestrians. Mean Walking Speed and Speed Variability also ranked highly, suggesting that faster and more inconsistent walking behaviors contributed significantly to model predictions. Stop-and-Go Duration further reinforced the relationship between motion irregularity and chaos by highlighting the role of discontinuous movement. Contextual features such as vehicle shares, traffic flow, and pedestrian density had comparatively lower SHAP values, indicating that daytime pedestrian chaos was largely governed by personal movement traits rather than environmental context.

For nighttime conditions (**Figure 3**), the SHAP results showed an even stronger dependence on individual-level features. Travel Duration, Travel Distance, and Speed Variability were the top three contributors, all positively correlated with predicted chaos. The association between longer and more variable trajectories and chaotic behavior may be amplified at night due to limited visibility, less structured pedestrian flow, and greater uncertainty. Path Efficiency and Stop-and-Go Duration also played a meaningful role. Pedestrians with low path efficiency or more frequent stops were more likely to exhibit higher motion irregularity, consistent with field observations. Vehicle-related variables and group-level features again showed minimal influence, reinforcing the notion that nighttime chaos is primarily shaped by internal pedestrian dynamics.

## CONCLUSIONS

This study presents a novel, data-driven framework for modeling chaotic pedestrian behavior using empirically observed trajectory data from a mixed-traffic highway environment. Given the need to capture real-world variability in pedestrian behavior, this research advances beyond traditional models that assume regularity or depend solely on simulation. While prior work has acknowledged the presence of chaos in pedestrian dynamics, most efforts have remained descriptive or theoretical, without advancing predictive tools that can quantify and anticipate such behavior using observable features. The objective was to bridge this gap by creating a pipeline that not only quantifies chaos using established nonlinear dynamics metrics but also leverages supervised machine learning to model this complexity from measurable inputs. By doing so, this study contributes to a scalable and interpretable approach to understanding micro-level pedestrian instability in real-world settings.

This study modeled chaotic pedestrian behavior using real-world trajectory data enriched with motion, group, and traffic features. Four chaos metrics were computed and combined using PCA. For both daytime and nighttime, CatBoost regressors trained on the PCA score were selected as the best-performing models.

By modeling behavioral chaos as a continuous, interpretable outcome grounded in observable features, this approach provides transportation planners, safety engineers, and simulation developers with a new tool for quantifying and anticipating pedestrian instability. Urban planners can use these models to flag high-risk crossing zones or poorly designed spaces where unpredictable behavior is likely to emerge, informing targeted interventions such as signage, barriers, or redesigned pedestrian infrastructure. For simulation platforms and digital twins, chaos-informed parameters can enhance realism by injecting behavioral noise reflective of actual human variability. In the realm of safety analytics, chaos scores can serve as early-warning signals for erratic movement patterns, particularly in shared spaces or areas with poor visibility. Moreover, connected and automated vehicle systems stand to benefit from understanding chaotic pedestrian behavior, as doing so enables more adaptive, risk-aware control algorithms when navigating through pedestrian-rich environments.

## ACKNOWLEDGEMENT

This research received no external funding. Grammarly was used for grammar and writing standards.

## AUTHOR CONTRIBUTIONS

The authors confirm contribution to the paper as follows: study conception and design: M. M. Shahrier, N. Haque, M. A. Raihan, M. Hadiuzzaman; data collection: M. M. Shahrier; analysis and interpretation of results: M. M. Shahrier, N. Haque, M. A. Raihan, M. Hadiuzzaman; draft manuscript preparation: M. M. Shahrier, N. Haque, M. A. Raihan. All authors reviewed the results and approved the final version of the manuscript.